%% file: 00_main.tex
\documentclass[10pt,twocolumn,letterpaper]{article}

\usepackage{cvpr}              
\usepackage{multirow}
\usepackage{rotating}
\usepackage{booktabs}
\usepackage{multirow}
\usepackage{tabularx}
\usepackage{booktabs}
\usepackage{tabularray}
\usepackage{graphics}
\usepackage{amssymb}
\usepackage{pifont}
\usepackage{lipsum}
\usepackage[title]{appendix}
\usepackage[accsupp]{axessibility}

\input{preamble}

\definecolor{cvprblue}{rgb}{0.21,0.49,0.74}
\usepackage[pagebackref,breaklinks,colorlinks,citecolor=cvprblue]{hyperref}

\def\paperID{6095} 
\def\confName{CVPR}
\def\confYear{2024}

\title{\model{}: A Single  Model for 2D and 3D Segmentation }

\author{
		Ayush Jain$^{1}$, Pushkal Katara$^{1}$, Nikolaos Gkanatsios$^{1}$, Adam W. Harley$^{2}$,\\ Gabriel Sarch$^{1}$, Kriti Aggarwal$^{3}$, Vishrav Chaudhary$^{3}$, Katerina Fragkiadaki$^1$\\
		$^1$Carnegie Mellon University, $^2$Stanford University, $^3$ Microsoft
		\\
		{\tt \small{\{ayushj2, pkatara ngkanats,gsarch,kfragki2\}@andrew.cmu.edu}}\\
   {\tt \small{aharley@cs.stanford.edu, \{kragga,vchaudhary\}@microsoft.com}}
	}
 
\begin{document}
 \maketitle


\input{0_abstract}    
 \input{1_introduction}

 \input{2_related}

 \input{3_methods}
 \input{4_experiments}
 \input{5_conclusion}
\input{8_acknowledgements}

 \begin{appendices}
     \input{6_supp_full}
 \end{appendices}

\clearpage
{
    \small
    \bibliographystyle{ieeenat_fullname}
    \bibliography{main}
}

\end{document}

%% file: preamble.tex
\usepackage[dvipsnames]{xcolor}

\newcommand{\model}{ODIN}

\definecolor{three_d_color}{HTML}{dab543}
\definecolor{two_d_color}{HTML}{087ca6}


%% file: 0_abstract.tex
\begin{abstract}
State-of-the-art models on contemporary 3D segmentation benchmarks like ScanNet
consume and label dataset-provided 3D point clouds, obtained through post processing of sensed multiview RGB-D images.
They are typically trained in-domain, forego  large-scale 2D pre-training and outperform alternatives that featurize the posed RGB-D multiview  images instead.  
The gap in performance between methods that consume posed images versus post-processed 3D point clouds   
has fueled the belief that 2D and 3D perception  require distinct model architectures. In this paper, we challenge this view and  
 propose \model{} (Omni-Dimensional INstance segmentation), 
 a  model that can segment and label both  2D RGB images  and  3D point clouds, using a transformer architecture that 
 alternates between 2D within-view  and 3D cross-view information fusion. 
 Our model differentiates 2D and 3D  feature operations    through the positional encodings of the tokens involved, which capture pixel coordinates for 2D patch tokens
 and  3D coordinates for 3D feature tokens.
 \model{} achieves state-of-the-art performance on  ScanNet200, Matterport3D and AI2THOR 3D  instance segmentation benchmarks, and competitive performance on ScanNet, S3DIS and  COCO. 
 It outperforms all previous works by a wide margin when the sensed 3D point cloud is used in place of the point cloud sampled from 3D mesh. 
 When used as the 3D perception engine in an instructable embodied agent architecture, it sets a new state-of-the-art on the TEACh action-from-dialogue benchmark. Our code and checkpoints can be found at the project website  \url{https://odin-seg.github.io}.  

\end{abstract}

%% file: 1_introduction.tex
\section{Introduction}

\begin{figure*}[!t]
\centering
    \includegraphics[width=\textwidth]{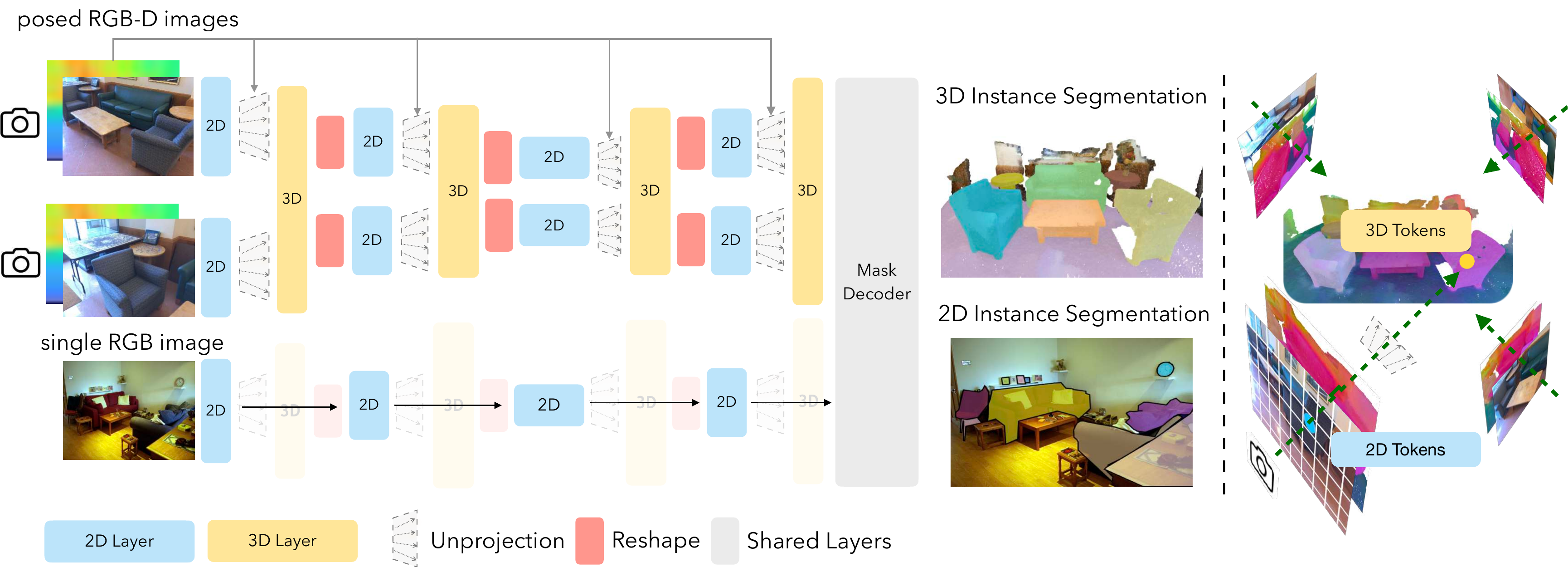}
    \caption{\textbf{Omni-Dimensional INstance segmentation (\model{})} is a model that can parse either a single RGB image or a multiview posed RGB-D sequence into 2D or 3D labelled object segments respectively. \textbf{Left}: Given a posed RGB-D sequence as input, \model{} alternates between a within-view 2D fusion and a cross-view 3D fusion. When the input is a single RGB image, the 3D fusion layers are skipped. \model{} shares the majority of its parameters across both RGB and RGB-D inputs, enabling the use of pre-trained 2D backbones. \textbf{Right}: At each 2D-to-3D transition, \model{} unprojects 2D feature tokens to their 3D locations using  sensed depth and camera intrinsics and extrinsics.}
    \label{fig:teaser}
\end{figure*}


There has been a surge of interest in porting 2D foundational image features to 3D scene understanding~\cite{conceptfusion,openscene,openmask3d,semabs,vlfield,pla,lerf,pl,deepview}. 
Some methods  lift  pre-trained 2D image features using sensed depth to 3D feature clouds~\cite{openscene,openmask3d,deepview,pla}. Others distill 2D backbones to  differentiable parametric 3D models, e.g., NeRFs,  by training  them per scene to render 2D feature maps of pre-trained backbones~\cite{lerf,pl}.  
Despite this effort, and despite the ever-growing power of 2D backbones~\cite{odise,m2f}, the state-of-the-art 
on established 3D segmentation benchmarks such as ScanNet~\cite{scannet} and ScanNet200~\cite{scannet200} \textit{still} consists of models that 
operate directly in 3D, without any 2D pre-training stage~\cite{mask3d,maft}.
Given the obvious power of 2D pre-training, why is it so difficult to yield improvements in these 3D tasks? 

We observe that part of the issue lies in a key implementation detail underlying these 3D benchmark evaluations. 
Benchmarks like ScanNet do not actually ask methods to use RGB-D images as input, even though this is the sensor data. Instead, these benchmarks first register all RGB-D frames into a single colored point cloud and reconstruct the scene as cleanly as possible, relying on manually tuned  stages for bundle adjustment, outlier rejection and meshing, and ask models to label the \textit{output reconstruction}. 
While it is certainly viable to scan and reconstruct a room before labelling any of the objects inside, this pipeline is perhaps inconsistent with the goals of embodied vision (and typical 2D vision), which involves dealing with actual sensor data and accounting for missing or partial observations.  
We therefore hypothesize that method rankings will change, and the impact of 2D pre-training will become evident, if we force the 3D models to take posed RGB-D frames as input rather than pre-computed mesh reconstructions. 
Our revised evaluation setting also opens the door to new methods, which can train and perform inference in either single-view or multi-view settings, with either RGB or RGB-D sensors. 

We  propose \textbf{O}mni-\textbf{D}imensional \textbf{IN}stance segmentation (\model{})$^{\dagger}$\let\thefootnote\relax\footnotetext{$^{\dagger}$The Norse god Odin sacrificed one of his eyes for wisdom, trading one mode of perception for a more important one. Our approach sacrifices 
perception on post-processed meshes for perception on posed RGB-D images. 
}, a model for 2D and 3D object segmentation and labelling that can parse single-view RGB images and/or multiview posed RGB-D images. 
As shown in \cref{fig:teaser}, \model{} alternates between 2D and 3D stages 
in its architecture, fusing information in 2D within each  image view, and in 3D across posed image views. At each 2D-to-3D transition, it unprojects 2D tokens to their 3D locations using the depth maps and camera parameters, 
and at each 3D-to-2D transition, it projects 3D tokens back to their image locations. 
Our model differentiates between 2D and 3D features  through the positional encodings of the tokens involved, which capture pixel coordinates for 2D patch tokens and  3D coordinates for 3D feature tokens. When dealing with 2D single-view input, our architecture simply skips the 3D layers and makes a forward pass with 2D layers alone. 

We test \model{} in 2D and 3D instance segmentation and 3D semantic segmentation on the 2D COCO object segmentation benchmark and the 3D benchmarks of ScanNet~\cite{scannet}, ScanNet200~\cite{scannet200}, Matterport3D~\cite{matterport}, S3DIS~\cite{s3dis} and AI2THOR~\cite{ai2thor,procthor}. 
When compared to methods using pre-computed mesh point cloud as input, our approach performs slightly worse than state-of-the-art on ScanNet and S3DIS, but better on ScanNet200 and Matterport3D. When using real sensor data as input with poses obtained from bundle reconstruction for all methods, our method performs even better, outperforming all prior work by a wide margin, in all datasets. 
We demonstrate that our model's ability to  jointly train on 3D and 2D datasets results in performance increase on 3D benchmarks, and also yields competitive segmentation accuracy on the 2D COCO benchmark. 
Our ablations show that interleaving 2D and 3D fusion operations outperforms designs where we first process in 2D and then move to 3D, or simply paint 3D points with 2D features. 
Stepping toward our broader goal of embodied vision, we also deploy \model{} as the 3D object segmentor of a SOTA embodied agent model~\cite{helper} on the simulation benchmark 
TEACh~\cite{teach} in the setup with access to RGB-D and pose information from the simulator, and demonstrate that our model sets a new state-of-the-art.
We make our code publicly available at
\url{https://odin-seg.github.io}.

%% file: 2_related.tex
\section{Related Work}

\paragraph{3D Instance Segmentation} 
Early methods in 3D instance segmentation~\cite{pointgroup,occuseg,hais,sstnet,softgroup,pbnet} group their semantic segmentation outputs into individual instances. Recently, Mask2Former~\cite{m2f} achieved state-of-the-art in 2D instance segmentation by instantiating \textit{object queries}, each directly predicting an instance segmentation mask by doing dot-product with the feature map of the input image. Inspired by it, Mask3D~\cite{mask3d} abandons the grouping strategy of prior 3D models to use the simple decoder head of Mask2Former. 
MAFT~\cite{maft} and QueryFormer~\cite{queryformer} improve over Mask3D by incorporating better query initialization strategies and/or relative positional embeddings. 
While this shift to Mask2Former-like architecture brought the 3D instance segmentation architectures closer to their 2D counterparts, the inputs and backbones remain very different: 2D models use pre-trained backbones~\cite{resnet,swin}, while 3D methods~\cite{mask3d} operate over point clouds and use sparse convolution-based backbones \cite{minkowski}, trained from scratch on small-scale 3D datasets. In this work, we propose to directly use RGB-D input and design architectures that can leverage strong 2D backbones to achieve strong performance on 3D benchmarks.

\paragraph{3D Datasets and Benchmarks} Most 3D models primarily operate on point clouds, avoiding the use of image-based features partly due to the design of popular benchmarks. These benchmarks generate point clouds by processing raw RGB-D sensor data, involving manual and noisy steps that result in misalignments between the reconstructed point cloud and sensor data. For instance, ScanNet~\cite{scannet} undergoes complex mesh reconstruction steps, including bundle reconstruction, implicit TSDF representation fitting, marching cubes, merging and deleting noisy mesh vertices, and finally manual removal of mesh reconstruction with high misalignments. Misalignments introduced by the mesh reconstruction process can cause methods processing sensor data directly to underperform compared to those trained and tested on provided point clouds. 
Additionally, some datasets, like HM3D~\cite{hm3d} lack access to raw RGB-D data. While mesh reconstruction has its applications, many real-time applications need to directly process sensor data. 
\vspace{-1em}

\paragraph{2D-based 3D segmentation}
Unlike instance segmentation literature, several approaches for semantic segmentation like MVPNet~\cite{mvpnet}, BPNet~\cite{bpnet} and DeepViewAgg~\cite{deepview} utilize the sensor point cloud directly instead of the mesh-sampled point cloud. Virtual Multiview Fusion~\cite{virtualview} forgoes sensor RGB-D images in favour of rendering RGB-D images from the provided mesh to fight misalignments and low field-of-view in ScanNet images.  Similar to our approach, BPNet and DeepViewAgg integrate 2D-3D information at various feature scales and initialize their 2D streams with pre-trained features. Specifically, they employ separate 2D and 3D U-Nets for processing the respective modalities and fuse features from the two streams through a connection module. Rather than employing distinct streams for featurizing raw data, our architecture instantiates a single unified U-Net which interleaves 2D and 3D layers and can handle both 2D and 3D perception tasks with a single unified architecture. 
Notably, while these works focus solely on semantic segmentation, our single architecture excels in both semantic and instance segmentation tasks. 

Recent advancements in 2D foundation models~\cite{clip,sam} have spurred efforts to apply them to 3D tasks such as point cloud classification~\cite{pix4point,pointclip,image2point}, zero-shot 3D semantic segmentation~\cite{openscene,conceptfusion,semabs} and more recently, zero-shot instance segmentation~\cite{openmask3d}. Commonly, these methods leverage 2D foundation models to featurize RGB images, project 3D point clouds onto these images, employ occlusion reasoning using depth and integrate features from all views through simple techniques like mean-pooling. Notably, these approaches predominantly focus on semantic segmentation, emphasizing pixel-wise labeling, rather than instance labeling, which necessitates  cross-view reasoning to associate the same object instance across multiple views. OpenMask3D~\cite{openmask3d} is the only method that we are aware of that attempts 3D instance segmentation using 2D foundation models, by training a class-agnostic 3D object segmentor on 3D point clouds and labelling it utilizing CLIP features. Despite their effectiveness in a zero-shot setting, they generally lag behind SOTA 3D supervised methods by 15-20\%. Rather than relying on features from foundation models, certain works~\cite{2d3d,pseudo} create 3D pseudo-labels using pre-trained 2D models. Another line of work involves fitting Neural-Radiance Fields (NeRFs), incorporating features from CLIP~\cite{lerf,vlfield} or per-view instance segmentations from state-of-the-art 2D segmentors~\cite{pl}. These approaches require expensive per-scene optimization that prohibits testing on all test scenes to compare against SOTA 3D discriminative models.
Instead of repurposing 2D foundation models for 3D tasks, Omnivore~\cite{omnivore} proposes to build a unified architecture that can handle multiple visual modalities like images, videos and single-view RGB-D image but they only show results for classification tasks. We similarly propose a single unified model capable of performing both single-view 2D and multi-view 3D instance and semantic segmentation tasks while utilizing pre-trained weights for the majority of our architecture. 

%% file: 3_methods.tex
\begin{figure*}[!ht]
     \centering
     \includegraphics[width=\textwidth]{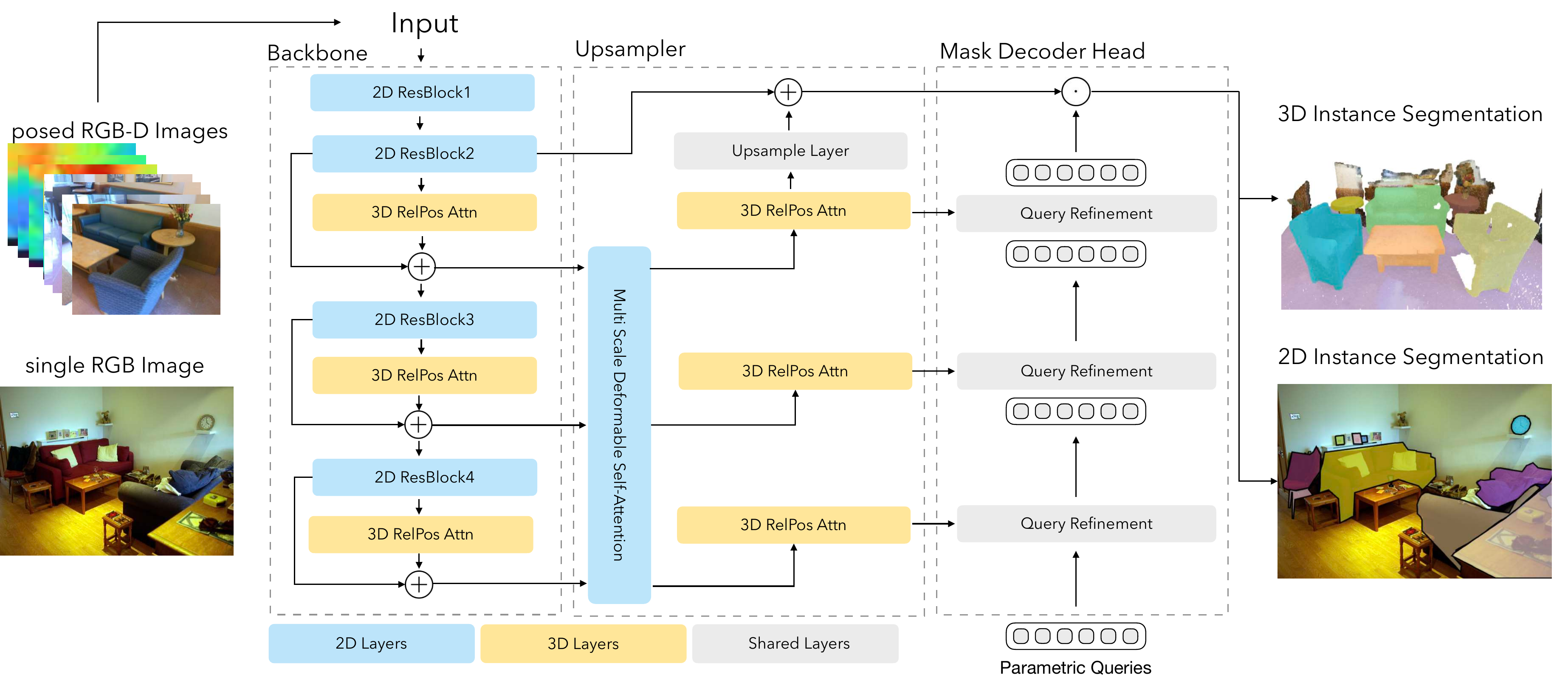}
    \caption{ \textbf{\model{} Architecture}: The input to our model is either a single RGB image or a multiview RGB-D posed sequence. We feed them to \model{}'s backbone which interleaves 2D within-view fusion layers and 3D cross-view attention layers to extract feature maps of different resolutions (scales). These feature maps exchange information through a multi-scale attention operation. Additional 3D fusion layers are used to improve multiview consistency. Then, a mask decoder head is used to initialize and refine learnable slots that attend to the multi-scale feature maps and predict object segments (masks and semantic classes).}
     \label{fig:method}
\end{figure*}


\section{Method}
\model{}'s architecture is shown in \cref{fig:method}. It takes either a single RGB image or a set of posed RGB-D images (i.e., RGB images associated with depth maps and camera parameters) and outputs the corresponding 2D or 3D instance segmentation masks and their semantic labels.
To achieve this, \model{} alternates between a 2D within-view fusion and a 3D attention-based cross-view fusion, as illustrated in \textcolor{two_d_color}{blue blocks} and  \textcolor{three_d_color}{yellow blocks} in  \cref{fig:method}. A segmentation decoding head predicts instance masks and semantic labels. Notably, \model{} shares the majority of its parameters across both RGB and multiview RGB-D inputs. We detail the components of our architecture below.

\noindent \textbf{Within-view 2D fusion:} We start from a 2D backbone, such as  ResNet50~\cite{resnet}  or Swin  Transformer~\cite{swin}, pre-trained for 2D COCO instance segmentation following  Mask2Former~\cite{m2f}, a state-of-the-art 2D segmentation model. 
When only a single RGB image is available, 
we pass it through the full backbone to obtain 2D features at multiple scales. When a posed RGB-D sequence is available, this 2D processing is interleaved with 3D stages, described next. 
By interleaving within-view and cross-view contextualization, we are able to utilize the pre-trained features from the 2D backbone while also fusing features across views, making them 3D-consistent. 

\noindent \textbf{Cross-view 3D fusion:} 
The goal of cross-view fusion is to make the individual images' representations consistent across views. 
As we show in our ablations, cross-view feature consistency is essential for 3D instance segmentation: it enables the segmentation head to realize that a 3D object observed from multiple views is indeed a single instance, rather than a separate instance in each viewpoint. 

\noindent \textit{1. 2D-to-3D Unprojection:} We unproject each  2D feature map to 3D by lifting each feature vector to a corresponding 3D location, using nearest neighbor depth and known camera intrinsic and extrinsic parameters, using a pinhole camera model. Subsequently, the resulting featurized point cloud undergoes voxelization, where the 3D space is discretized into a volumetric grid. 
Within each occupied grid cell (voxel), the features and XYZ coordinates are mean-pooled to derive new sets of 3D feature tokens and their respective 3D locations.


\noindent \textit{2. 3D $k$-NN Transformer with Relative Positions:} We fuse information across 3D tokens using $k$-nearest-neighbor attention with relative 3D positional embeddings. This is similar to Point Transformers~\cite{ptv1, ptv2}, but we simply use vanilla cross-attention instead of the vector attention proposed in those works. 
Specifically, in our approach, each 3D token attends to its $k$ nearest neighbors. The positional embeddings in this operation are relative to the query token's location. We achieve this by encoding the distance vector between a token and its neighbour with an MLP. The positional embedding for the query is simply encoding of the $0$ vector. We therefore have 
\begin{align}
    \textrm{query}_{\textrm{pos}} &= \textrm{MLP}(0); \\ 
    \textrm{key}_{\textrm{pos}} &= \textrm{MLP}(p_i - p_j),
\end{align} 
where $p_i$ represents the 3D tokens, shaped $N \times 1 \times 3$, and  $p_j$ represents the $k$ nearest neighbors of each $p_i$, shaped $N \times k \times 3$. 
In this way, the attention operation is invariant to the absolute coordinates of the 3D tokens and only depends on their relative spatial arrangements. 
While each 3D token always attends to the same $k$ neighbors, its effective receptive field grows across layers, as the neighbors' features get updated when they perform their own attention~\cite{graph}. 

\noindent \textit{3. 3D-to-2D Projection:} After contextualizing the tokens in 3D, we project the features back to their original 2D locations. We first copy the feature of each voxel to all points within that voxel. We then reshape these points back into multiview 2D feature maps, so that they may be processed by the next 2D module. 
The features vectors are unchanged in this transition; the difference lies in their interpretation and shape. In 2D the features are shaped $V \times H \times W \times F$, representing a feature map for each viewpoint, and in 3D they are shaped $N \times F$, representing a unified feature cloud, where $N = V\cdot H \cdot W$. 


\noindent \textbf{Cross-scale fusion and upsampling:}
After multiple single-view and cross-view stages, we have access to multiple features maps per image, at different resolutions. 
We merge these with the help of deformable 2D attention, 
akin to Mask2Former \cite{m2f}, operating on the three lowest-resolution scales $(1/32, 1/16, 1/8)$. 
When we have 3D input, we apply an additional 3D fusion layer at each scale after the deformable attention, to restore the 3D consistency.  
Finally, we use a simple upsampling layer on the $1/8$ resolution feature map to bring it to $1/4$ resolution and add with a skip connection to the $1/4$ feature map from the backbone. 


\noindent \textbf{Sensor depth to mesh point cloud feature transfer:}
For 3D benchmarks like ScanNet~\cite{scannet} and ScanNet200~\cite{scannet200}, the objective is to label a point cloud derived from a mesh rather than the depth map from the sensor. Hence, on those benchmarks, instead of upsampling the $1/8$ resolution feature map to $1/4$, we trilinearly interpolate features from the  $1/8$ resolution feature map to the provided point cloud sampled from the mesh. This means: for each vertex in the mesh, we trilinearly interpolate from our computed 3D features to obtain interpolated features. We additionally similarly interpolate from the unprojected $1/4$ resolution feature map in the backbone, for an additive skip connection. 



\noindent \textbf{Shared 2D-3D segmentation mask decoder:} Our segmentation decoder  is a Transformer, similar to Mask2Former's decoder head,  which takes as input upsampled 2D or 3D feature maps and outputs corresponding 2D or 3D  segmentation masks and their semantic classes. 
Specifically, we instantiate a set of $N$ learnable object queries responsible for decoding individual instances. These queries are iteratively refined by
a \textit{Query Refinement} block, which consists of cross-attention to the upsampled features, followed by a self-attention between the queries. Except for the positional embeddings, all attention and query weights are shared between 2D and 3D. We use Fourier positional encodings in 2D, while in 3D we encode the XYZ coordinates of the 3D tokens with an MLP. The refined queries are used to predict instance masks and semantic classes. For mask prediction, the queries do a token-wise dot product with the highest-resolution upsampled features. For semantic class prediction, we use an MLP over the queries, mapping them to class logits. We refer readers to Mask2Former~\cite{m2f} for further details.

\noindent \textbf{Open vocabulary class decoder:} Drawing inspiration from prior open-vocabulary detection methods~\cite{butd,glip,xdecoder}, we introduce an alternative classification head capable of handling an arbitrary number of semantic classes. This modification is essential for joint training on multiple datasets.
Similar to BUTD-DETR~\cite{butd} and GLIP~\cite{glip}, we supply the model with a \textit{detection prompt} formed by concatenating object categories into a sentence (e.g., \textit{``Chair. Table. Sofa."}) and encode it using RoBERTa~\cite{roberta}. In the query-refinement block, queries additionally attend to these text tokens before attending to the upsampled feature maps. 
For semantic class prediction, we first perform a dot-product operation between queries and language tokens, generating one logit per token in the detection prompt. The logits corresponding to prompt tokens for a specific object class are then averaged to derive per-class logits. This can handle multi-word noun phrases such as \textit{``shower curtain"}, where we average the logits corresponding to \textit{``shower"} and \textit{``curtain"}. 
The segmentation masks are predicted by a pixel-/point-wise dot-product, in the same fashion as described earlier.

\noindent \textbf{Implementation details:} 
We initialize our model with pre-trained weights from Mask2Former~\cite{m2f} trained on COCO~\cite{coco}. Subsequently, we train all parameters end-to-end, including both pre-trained and new parameters from 3D fusion layers. During training in 3D scenes, our model processes a sequence of $N$ consecutive frames, usually comprising 25 frames. At test time, we input all images in the scene to our model, with an average of 90 images per scene in ScanNet. We use vanilla closed-vocabulary decoding head for all experiments except when training jointly on 2D-3D datasets. There we use our open vocabulary class decoder that lets us handle different label spaces in these datasets. During training, we employ open vocabulary mask decoding for joint 2D and 3D datasets and vanilla closed-vocabulary decoding otherwise.
Training continues until convergence on 2 NVIDIA A100s with 40 GB VRAM, with an effective batch size of 6 in 3D and 16 in 2D. For joint training on 2D and 3D datasets, we alternate sampling 2D and 3D batches with batch sizes of 3 and 8 per GPU, respectively. We adopt Mask2Former's strategy, using Hungarian matching for matching queries to ground truth instances and supervision losses.
While our model is only trained for instance segmentation, it can perform semantic segmentation for free at test time like Mask2Former. We refer to Mask2Former~\cite{m2f} for more details.

%% file: 4_experiments.tex
\section{Experiments}

\subsection{Evaluation on  3D benchmarks}
\label{sec:real_3d}
\input{tables/real_in_domain_3d}
\noindent \textbf{Datasets:} First, we test our model on 3D instance and semantic segmentation in   the ScanNet~\cite{scannet} and ScanNet200~\cite{scannet200} benchmarks. The goal of these benchmarks is to label the point cloud extracted from the 3D mesh of a scene reconstructed from raw sensor data. ScanNet evaluates on 20 common semantic classes, while ScanNet200 uses 200 classes, which is more representative of the long-tailed object distribution encountered in the real world. We report results on the official validation split of these datasets here and on the official test split in the supplementary.

\noindent \textbf{Evaluation metrics:} We follow the standard evaluation metrics, namely mean Average Precision (mAP) for instance segmentation and mean Intersection over Union (mIoU) for semantic segmentation.

\noindent \textbf{Baselines:} In \textit{instance segmentation}, our main baseline is the SOTA 3D method Mask3D~\cite{mask3d}. For a thorough comparison, we train both  Mask3D and our model with sensor RGB-D point cloud input and evaluate them on the benchmark-provided mesh-sampled point clouds. We also compare with the following recent and concurrent works: PBNet~\cite{pbnet}, QueryFormer~\cite{queryformer} and MAFT~\cite{maft}. QueryFormer and MAFT explore  query initialization and refinement in a Mask3D-like architecture and thus have complementary advantages to ours. Unlike \model{}, these methods directly process 3D point clouds and initialize their weights from scratch.
 As motivated before, utilizing RGB-D input directly has several advantages, including avoiding costly mesh building processes, achieving closer integration of 2D and 3D perception, and leveraging pre-trained features and abundant 2D data. 

In \textit{semantic segmentation}, we compare with MVPNet~\cite{mvpnet}, BPNet~\cite{bpnet} and state-of-the-art DeepViewAgg~\cite{deepview} which directly operate on sensor RGB or RGB-D images and point clouds. We also compare with VMVF~\cite{virtualview} which operates over rendered RGB-D images from the provided mesh, with heuristics for camera view sampling to avoid occlusions, ensures balanced scene coverage, and employs a wider field-of-view, though we note their code is not publicly available. 
Similar to \model{}, all of these methods utilize 2D pre-trained backbones. We also compare with Point-Transformer v2 \cite{ptv2}, Stratified Transformer \cite{stratified}, OctFormer \cite{octformer} and Swin3D-L \cite{swin3d} which process the mesh-sampled point cloud directly, without using any 2D pre-training.  On the ScanNet200 semantic segmentation benchmark, we compare with SOTA OctFormer \cite{octformer} and with CeCo~\cite{ceco}, a method specially designed to fight class-imbalance in ScanNet200. These methods directly process the point cloud and do not use 2D image pre-trained weights. We also compare with LGround \cite{scannet200} which uses 2D CLIP pre-training. We also compare with zero-shot 2D foundation model-based 3D models of OpenScene~\cite{openscene} and OpenMask3D~\cite{openmask3d}. This comparison is unfair since they are not supervised within-domain, but we include them for completeness.
The results are presented in \cref{tab:established_3d}. 
We draw the following conclusions: 


\noindent \textbf{Performance drops with sensor point cloud as input} (\cref{tab:scannet_ins}):     Mask3D's performance drops from 55.2\% mAP with mesh point cloud input to 43.9\% mAP with sensor point cloud input. This is consistent with prior works~\cite{virtualview,deepview} in 3D semantic segmentation on ScanNet, which attributes the drop to misalignments caused by noise in camera poses, depth variations and post-processing steps.

\noindent \textbf{\model{} outperforms SOTA 3D methods with sensor point cloud input and underperforms them when baselines use mesh-sampled point clouds} (\cref{tab:scannet_ins}): Our model significantly outperforms SOTA Mask3D model with sensor point cloud input and achieves comparable performance to methods using mesh-sampled point cloud input on the mAP25 metric while far behind on mAP metric, due to misalignments between the 3D mesh and the sensor point cloud.
    
\noindent \textbf{\model{} sets a new SOTA in semantic segmentation on ScanNet} (\cref{tab:scannet_sem}) outperforming all methods on all setups including the models trained on the sensor, rendered and mesh sampled point clouds. 



\noindent \textbf{\model{} sets a new instance segmentation SOTA on the long-tailed ScanNet200 dataset} (\cref{tab:scannet200_ins}) outperforming SOTA 3D models on all setups including the models trained on mesh-sampled point cloud, especially by a large margin in mAP25 metric, while exclusively utilizing sensor RGB-D data. This highlights the contribution of 2D features, particularly in detecting a long tail of class distribution where limited 3D data is available. We show more detailed results with performance on the head, common and tail classes in the appendix (~\cref{subsec:scannet200_detail}).


\noindent \textbf{\model{} sets a new semantic segmentation SOTA on ScanNet200} (\cref{tab:scannet200_sem}), outperforming SOTA semantic segmentation models that use mesh point clouds.
    
    

\input{tables/ai2thor}

\subsection{Evaluation on multiview RGB-D in simulation}
Using the AI2THOR~\cite{ai2thor} simulation environment with procedural homes from ProcThor~\cite{procthor}, we collected RGB-D data for 1500 scenes (1200 training, 300 test) of similar size as ScanNet (more details in appendix, ~\cref{subsec:impl_detail}).
We train and evaluate our model and SOTA Mask3D~\cite{mask3d} on the unprojected RGB-D images. As shown in \cref{tab:sim}, our model outperforms Mask3D by 3.7\% mAP, showing strong performance in a directly comparable RGB-D setup.
It suggests that current real-world benchmarks may restrain models that featurizes RGB-D sensor point clouds due to misalignments. We hope this encourages the community to also focus on directly collecting, labeling, and benchmarking RGB-D sensor data. 



\subsection{Embodied Instruction Following}
\input{tables/teach}
\input{tables/joint_2d_3d}

\input{tables/ablations_alt}

We apply \model{} in the embodied setups of TEACh~\cite{teach} and ALFRED~\cite{alfred} where agents have access to RGB, depth and camera poses and need to infer and execute task and action plans from dialogue segments and instructions, respectively. These agents operate in dynamic home environments and cannot afford expensive mesh building steps. 
Detecting objects well is critical for task success in both cases. Prior SOTA methods~\cite{helper,teach} run per-view 2D instance segmentation models \cite{m2f,solq} and link the detected instances using simple temporal reasoning regarding spatial and appearance proximity. Instead, \model{} processes its last $N$ egocentric views and segments objects instances directly in 3D. We equip HELPER \cite{helper}, a state-of-the-art embodied model, with \model{} as its 3D object detection engine. We evaluate using Task Sucess Rate (SR) which checks if the entire task is executed successfully, and Goal Conditioned Success Rate (GC) which checks the proportion of satisfied subgoals across all episodes~\cite{alfred,teach}. We perform evaluation on "valid-seen" (houses similar to the training set) and "valid-unseen" (dissimilar houses) splits. In \cref{tab:embodied}, we observe that HELPER with \model{} as its 3D object detector  significantly outperforms HELPER that uses the original 2D detection plus linking perception pipeline.

\subsection{Ablations and Variants}
We conduct our ablation experiments on the ScanNet dataset in  \cref{tab:joint} and \cref{tab:ablations}. Our conclusions are:

\noindent \textbf{Joint 2D-3D training helps 3D perception} We compare joint training of \model{} on sensor RGB-D point clouds from ScanNet and 2D RGB images from COCO to variants trained independently on 2D and 3D data, all initialized  from pre-trained COCO weights. Since there are different classes in ScanNet and COCO, we use our open-vocabulary semantic class-decoding head instead of the vanilla closed-vocabulary head. Results in \cref{tab:joint} show that joint training yields a 1.3\% absolute improvement in 3D, and causes a similar drop in 2D. This experiment indicates that a single architecture can perform well on both 2D and 3D tasks, thus indicating that we may not need to design vastly different architectures in either domain. However, the drop in 2D performance indicates a potential for further improvements in the architecture design to retain the performance in the 2D domain.  
Nevertheless, this experiment highlights the  benefits of joint training with 2D datasets for 3D segmentation in \model{}. 
Note that we do not jointly train on 2D and 3D datasets in any of our other experiments due to computational constraints.

\noindent \textbf{Cross-View fusion is crucial for instance segmentation but not for semantic segmentation} (\cref{tab:cross_view}): removing 3D cross-view fusion layers results in an 8.5\% mAP drop for instance segmentation, without any significant effect in semantic segmentation. Popular 2D-based 3D open vocabulary works~\cite{conceptfusion,openscene} without strong cross-view fusion only focus on semantic segmentation and thus could not uncover this issue. Row-3 shows a 6.1\% mAP drop when cross-view 3D fusion happens after all within-view 2D layers instead of interleaving the within-view and cross-view fusion. 


\noindent \textbf{2D pre-trained weight initialization helps} (\cref{tab:pretrain}): initializing only the image backbone with pre-trained weights, instead of all layers (except the 3D fusion layers), results in a 5.5\% mAP drop (row-2). Starting the entire model from scratch leads to a larger drop of 6.3\% mAP (row-3). This underscores the importance of sharing as many parameters as possible with the 2D models to leverage the maximum possible 2D pre-trained weights.

\noindent \textbf{Stronger 2D backbones helps} (\cref{tab:backbone_frozen}): using Swin-B over ResNet50 leads to significant performance gains, suggesting that \model{} can directly benefit from advancements in 2D computer vision. 

\noindent \textbf{Finetuning everything including the pre-trained parameters helps} (\cref{tab:backbone_frozen}): ResNet50's and Swin's performance increases substantially when we fine-tune all parameters. Intuitively, unfreezing the backbone allows 2D layers to adapt to cross-view fused features better. Thus, we keep our backbone unfrozen in all experiments. 


\noindent \textbf{Supplying 2D features directly to 3D models does not help:} Concatenating 2D features with XYZ+RGB as input to Mask3D yields 53.8\% mAP performance, comparable to 53.3\%\let\thefootnote\relax\footnotetext{$^{\dagger}$We do not use the expensive DB-SCAN post-processing of Mask3D, and hence it gets 53.3\% mAP instead of 55.2\% as reported by their paper} of the baseline model with only XYZ+RGB as input.

\subsection{Additional Experiments}
We show evaluations on the hidden test set of ScanNet and ScanNet200 in ~\cref{subsec:test_set}, results and comparisons with baselines on S3DIS \cite{s3dis} and MatterPort3D \cite{matterport} datasets in ~\cref{subsec:m3d_s3dis} and performance gains in 2D perception with increasing context views in ~\cref{subsec:num_views}.

\subsection{Limitations}

Our experiments reveal the following limitations for \model{}: 
Firstly, like other top-performing 3D models, it depends on accurate depth and camera poses. Inaccurate depth or camera poses cause a sharp decrease in performance (similar to other 3D models, like Mask3D). 
In our future work, we aim to explore unifying depth and camera pose estimation with semantic scene parsing, thus making 3D models more resilient to noise. 
Secondly, in this paper, we limited our scope to exploring the design of a unified architecture without scaling up 3D learning by training on diverse 2D and 3D datasets jointly. We aim to explore this in future to achieve strong generalization to in-the-wild scenarios, akin to the current foundational 2D perception systems. Our results suggest a competition between 2D and 3D segmentation performance when training \model{} jointly on both modalities. Exploring ways to make 2D and 3D training more synergistic is a promising avenue for future work.

%% file: tables/real_in_domain_3d.tex
\begin{table*}

\caption{\textbf{Evaluation on 3D Benchmarks} ($^\mathsection$ = trained by us using official codebase).}
\label{tab:established_3d}
\begin{subtable}[t]{0.53\textwidth}
\small
\centering
\caption{\textbf{ScanNet Instance Segmentation Task.}}
\label{tab:scannet_ins}
\resizebox{\textwidth}{!}{%
\begin{tabular}{llccccc}
\midrule
& Model & mAP &  mAP50 & mAP25 \\ \midrule
\multirow{3}{1.9cm}{Sensor RGBD Point Cloud} & Mask3D$^\mathsection$ \cite{mask3d} & 43.9 & 60.0 & 69.9  \\
 & \model{}-ResNet50 (Ours) & 47.8 & 69.8  & \textbf{83.6}  \\
 & \model{}-Swin-B (Ours) & \textbf{50.0} & \textbf{71.0} & \textbf{83.6}  \\ \midrule
\multirow{5}{1.9cm}{Mesh Sampled Point Cloud} & SoftGroup \cite{softgroup} & 46.0 & 67.6 & 78.9 \\
& PBNet \cite{pbnet} & 54.3 & 70.5 & 78.9 \\
& Mask3D \cite{mask3d} & 55.2 & 73.7 & \textbf{83.5} \\
& QueryFormer \cite{queryformer} & 56.5 & 74.2 & 83.3 \\
& MAFT \cite{maft} & \textbf{58.4} & \textbf{75.9} & - \\ 
 \midrule
\end{tabular}
}
\end{subtable}
\hspace{\fill}
\begin{subtable}[t]{0.48\textwidth}
\small
\centering
\caption{\textbf{ScanNet Semantic Segmentation Task.}}
\label{tab:scannet_sem}
\begin{tabular}{llc} 
\midrule
& Model & mIoU \\ \midrule
\multirow{5}{2.2cm}{Sensor RGBD Point Cloud} 
& MVPNet \cite{mvpnet} & 68.3\\
& BPNet \cite{bpnet} & 69.7 \\
& DeepViewAgg \cite{deepview} & 71.0 \\
 & \model{}-ResNet50 (Ours) & 73.3  \\
 & \model{}-Swin-B (Ours) & \textbf{77.8} \\ \midrule
\multirow{1}{2.2cm}{Rendered RGBD Point Cloud} & VMVF \cite{virtualview} & \textbf{76.4} \\ \\\midrule
\multirow{4}{2.2cm}{Mesh Sampled Point Cloud} & Point Transformer v2 \cite{ptv2} & 75.4 \\ 
& Stratified Transformer \cite{stratified} & 74.3 \\
& OctFormer \cite{octformer} & 75.7 \\
& Swin3D-L \cite{swin3d} & \textbf{76.7}  \\
 \midrule
Zero-Shot & OpenScene \cite{openscene} & 54.2 \\
\midrule
\end{tabular} 
\end{subtable}
\bigskip \\
\begin{subtable}[t]{0.52\textwidth}
\vspace*{-0.7 cm}
\small
\centering
\caption{\textbf{ScanNet200 Instance Segmentation Task.}}
\label{tab:scannet200_ins}
\resizebox{\textwidth}{!}{
\begin{tabular}{llccc}
\midrule
 & Model  & mAP &  mAP50 & mAP25 \\ \midrule
\multirow{3}{1.9cm}{Sensor RGBD Point Cloud} & Mask3D \cite{mask3d} $^{\mathsection}$ & 15.5 & 21.4 & 24.3  \\
 & \model{}-ResNet50 (Ours) & 25.6 & 36.9 & 43.8  \\
 & \model{}-Swin-B (Ours) &  \textbf{31.5}        &  \textbf{45.3}         &  \textbf{53.1}   \\ \midrule
\multirow{3}{1.9cm}{Mesh Sampled Point Cloud} & Mask3D \cite{mask3d} & 27.4 & 37.0  & 42.3 \\
& QueryFormer \cite{queryformer} & 28.1 & 37.1 & \textbf{43.4} \\
& MAFT \cite{maft} & \textbf{29.2} & \textbf{38.2} & 43.3 \\ 
 \midrule
 Zero-Shot & OpenMask3D \cite{openmask3d} & 15.4 & 19.9 & 23.1 \\
 \midrule
\end{tabular}
}
\end{subtable}
\hspace{\fill}
\begin{subtable}[t]{0.48\textwidth}
\small
\centering
\caption{\textbf{ScanNet200 Semantic Segmentation Task.}}
\label{tab:scannet200_sem}

\begin{tabular}{llc}
\midrule
& Model  & mIoU \\ \midrule
\multirow{2}{2cm}{Sensor RGBD Point Cloud} & \model{}-ResNet50 (Ours) & 35.8 \\
 & \model{}-Swin-B (Ours) &  \textbf{40.5}  \\ \midrule
\multirow{3}{2cm}{Mesh Sampled Point Cloud} & LGround \cite{scannet200} & 28.9 \\
& CeCo \cite{ceco} & 32.0 \\
& Octformer \cite{octformer} & \textbf{32.6} \\
 \midrule
\end{tabular}
\end{subtable}

\end{table*}

%% file: tables/ai2thor.tex
\begin{table}
\small
\centering
\caption{\textbf{AI2THOR Semantic and Instance Segmentation.}}
\label{tab:sim}
\centering
\vspace{-1em}
\begin{tabular}{lcccc}
\midrule
Model & mAP &  mAP50 & mAP25 & mIoU \\ \midrule
 Mask3D \cite{mask3d} & 60.6 & 70.8 & 76.6 & -  \\
 \model{}-ResNet50 (Ours) & 63.8  & \textbf{73.8} & \textbf{80.2} & \textbf{71.5}  \\
 \model{}-Swin-B (Ours) & \textbf{64.3}  & 73.7 & 78.6 & 71.4  \\ \midrule
\end{tabular}
\end{table}

%% file: tables/teach.tex
\begin{table}[t] 
\begin{center}
\small
\setlength{\tabcolsep}{3pt} 
\caption{\textbf{Embodied Instruction Following.}
SR = success rate. GC = goal condition success rate.
}\label{tab:embodied}
\vspace{-1em}
\resizebox{\columnwidth}{!}{%
\begin{tabular}{@{}p{2.5cm}ccccccccc@{}}
\midrule
 & \multicolumn{4}{c}{\textbf{TEACh}} & \multicolumn{4}{c}{\textbf{ALFRED}} \\
 \cmidrule(lr){2-5} \cmidrule(lr){6-9}
 & \multicolumn{2}{c}{\textbf{Unseen}} & \multicolumn{2}{c}{\textbf{Seen}} & \multicolumn{2}{c}{\textbf{Unseen}} & \multicolumn{2}{c}{\textbf{Seen}} \\
  & \multicolumn{1}{c}{SR} & \multicolumn{1}{c}{GC} & \multicolumn{1}{c}{SR} & \multicolumn{1}{c}{GC} & \multicolumn{1}{c}{SR} & \multicolumn{1}{c}{GC} & \multicolumn{1}{c}{SR} & \multicolumn{1}{c}{GC} \\ 
 \midrule
FILM \cite{film} & - & - & - & -  & 30.7 & 42.9  & 26.6 & 38.2  \\
HELPER \cite{helper}~ & 15.8  & 14.5  & 11.6  & 19.4  & 37.4  & 55.0  & 26.8  & 41.2  \\
\textsc{HELPER + \model{} (Ours)}~ & \textbf{18.6}  & \textbf{ 18.6}  & \textbf{13.8} & \textbf{26.6}  & \textbf{47.7} & \textbf{61.6}  & \textbf{33.5}  & \textbf{47.1}   \\
 \hline
\end{tabular}
}
\end{center}
\end{table}

%% file: tables/joint_2d_3d.tex
\begin{table}
\small
\centering
\caption{\textbf{Joint Training on Sensor RGB-D point cloud from ScanNet and 2D RGB images from COCO.}}
\label{tab:joint}
\vspace{-1em}
\resizebox{\columnwidth}{!}{%
\begin{tabular}{p{3.0cm}cccc}
\midrule

                  & \multicolumn{3}{c}{\textbf{ScanNet}} &     \multicolumn{1}{c}{\textbf{COCO}} \\
\cmidrule(lr){2-4} \cmidrule(lr){5-5}
                        & mAP     & mAP50 & mAP25~ & mAP  \\
\midrule
Mask3D \cite{mask3d}                      &     43.9    &  60.0     &   69.9        & \ding{55}  \\
Mask2Former \cite{m2f}                 & \ding{55}      & \ding{55}   & \ding{55} & \textbf{43.7}    \\
\model{} (trained in 2D) &    \ding{55}     & \ding{55}      &   \ding{55}      &   43.6   \\
\model{} (trained in 3D) &    47.8     & 69.8      &   \textbf{83.6}      &   \ding{55}   \\
\model{} (trained jointly) &  \textbf{49.1}    &     \textbf{70.1}    &    83.1     &    41.2  \\
\midrule
\end{tabular}
}
\end{table}

%% file: tables/ablations_alt.tex
\begin{table*}[t]
\caption{\textbf{Ablations on ScanNet Dataset.}}
\label{tab:ablations}
\begin{subtable}[t]{0.28\textwidth}
\small
    \caption{\textbf{Cross-View Contextualization.}}
    \label{tab:cross_view}
    \begin{tabular}{p{2.2cm}cc}
    \midrule
        Model & mAP & mIoU \\ \midrule
        \model{} (Ours) & \textbf{47.8} & 73.3 \\
        No 3D Fusion & 39.3 & 73.2 \\ 
        No interleaving & 41.7  & \textbf{73.6} \\
        \midrule
    \end{tabular}
\end{subtable}
\begin{subtable}[t]{0.35\textwidth}
\small
    \caption{\textbf{Effect of Pre-Trained Features.}}
    \label{tab:pretrain}
    \begin{tabular}{p{3.0cm}cc}
    \midrule
        Model & mAP & mIoU \\ \midrule
        \model{} (Ours) & \textbf{47.8} & \textbf{73.3} \\
        Only pre-trained backbone & 42.3  & 72.9 \\
        No pre-trained features &  41.5 & 68.6 \\
        \midrule
    \end{tabular}
\end{subtable}
\begin{subtable}[t]{0.3\textwidth}
 \small
    \caption{\textbf{Effect of Freezing Backbone.}}
    \label{tab:backbone_frozen}
    \begin{tabular}{p{2.0cm}cccc}
    \midrule
        Model & \multicolumn{2}{c}{ResNet50} & \multicolumn{2}{c}{Swin-B} \\
        & mAP & mIoU & mAP & mIoU \\
        \midrule
        \model{} (Ours) & \textbf{47.8} & 73.3 & \textbf{50.0} & \textbf{77.8} \\
        With frozen backbone & 46.7  & \textbf{74.3} & 46.2 & 75.9 \\
        \midrule
    \end{tabular}
\end{subtable}
\end{table*}

%% file: 5_conclusion.tex
\section{Conclusion}

We presented \model{}, a model for 2D and 3D instance segmentation that can parse 2D images and 3D point clouds alike. \model{} represents both 2D images and 3D feature clouds as a set of tokens that differ in their positional encodings which represent 2D pixel coordinates for 2D tokens and 3D XYZ coordinates for 3D tokens. Our model alternates between within-image featurization and cross-view featurization. It achieves SOTA performance in ScanNet200 and AI2THOR instance segmentation benchmarks, outperforms all methods operating on sensor point clouds and achieves competent performance to methods operating over mesh-sampled pointcloud. 
Our experiments show that \model{} outperforms alternative models that simply augment 3D point cloud models with  2D image features as well as ablative versions of our model that do not alternate between   2D and 3D information fusion, do not co-train across 2D and 3D and do no pre-train the 2D backbone.  

%% file: 8_acknowledgements.tex
\section{Acknowledgements}

The authors express gratitude to Wen-Hsuan Chu, Mihir Prabhudesai, and Alexander Swerdlow for their valuable feedback on the early draft of this work. Special thanks to Tsung-Wei Ke for insightful discussions throughout the project. 
We thank the Microsoft Turing Team for providing us with GPU resources during the initial development phase of this project.  This work is supported by ONR award N00014-23-1-2415, Sony AI,  DARPA Machine Common Sense, an Amazon faculty award, and an NSF CAREER award.

%% file: 6_supp_full.tex


\section{Experiments}
\input{tables/test_real_in_domain_3d}
\subsection{Evaluations on ScanNet and ScanNet200 Hidden Test Sets}
\label{subsec:test_set}
We submit \model{} to official test benchmarks of ScanNet \cite{scannet} and ScanNet200 \cite{scannet200}. Following prior works, we train \model{} on a combination of train and validation scenes. Unlike some approaches that employ additional tricks like DBSCAN \cite{mask3d}, ensembling models \cite{stratified}, additional specialized augmentations \cite{ptv2}, additional pre-training on other datasets \cite{swin3d}, finer grid sizes \cite{octformer} and multiple forward passes through points belonging to the same voxel, our method avoid any such bells and whistles.  


The results are shown in \cref{tab:test_established_3d}. All conclusions from results on the validation set of these datasets as discussed in the main paper are applicable here. On the ScanNet benchmark, \model{} achieves close to SOTA performance on semantic segmentation and mAP25 metric of Instance Segmentation while being significantly worse on mAP metric due to misalignments between sensor and mesh sampled point clouds. On ScanNet200 benchmark, \model{} sets a new SOTA on semantic segmentation and mAP50/mAP25 metric of Instance Segmentation, while achieving close to SOTA performance on mAP metric. Notably \model{} is the first method that operates over sensor RGB-D data for instance segmentation and achieves competitive performance to models operating over mesh-sampled point clouds.

\input{7_s3dis_m3d}

\subsection{ScanNet200 Detailed Results}
\label{subsec:scannet200_detail}
\input{tables/scannet200_detailed}

ScanNet200 \cite{scannet200} categorizes its 200 object classes into three groups—\textit{Head}, \textit{Common}, and \textit{Tail}—each comprising 66, 68, and 66 categories, respectively. In \cref{tab:scannet200_detailed}, we provide a detailed breakdown of the ScanNet200 results across these splits. We observe that in comparison to SOTA Mask3D model trained on mesh-sampled point cloud, \model{} achieves lower performance on \textit{Head} classes, while significantly better performance on \textit{Common} and \textit{Tail} classes. This highlights the contribution of effectively utilizing 2D pre-trained features, particularly in detecting a long tail of class distribution where limited 3D data is available.

\begin{figure}[t]
     \centering
     \includegraphics[width=0.48\textwidth]{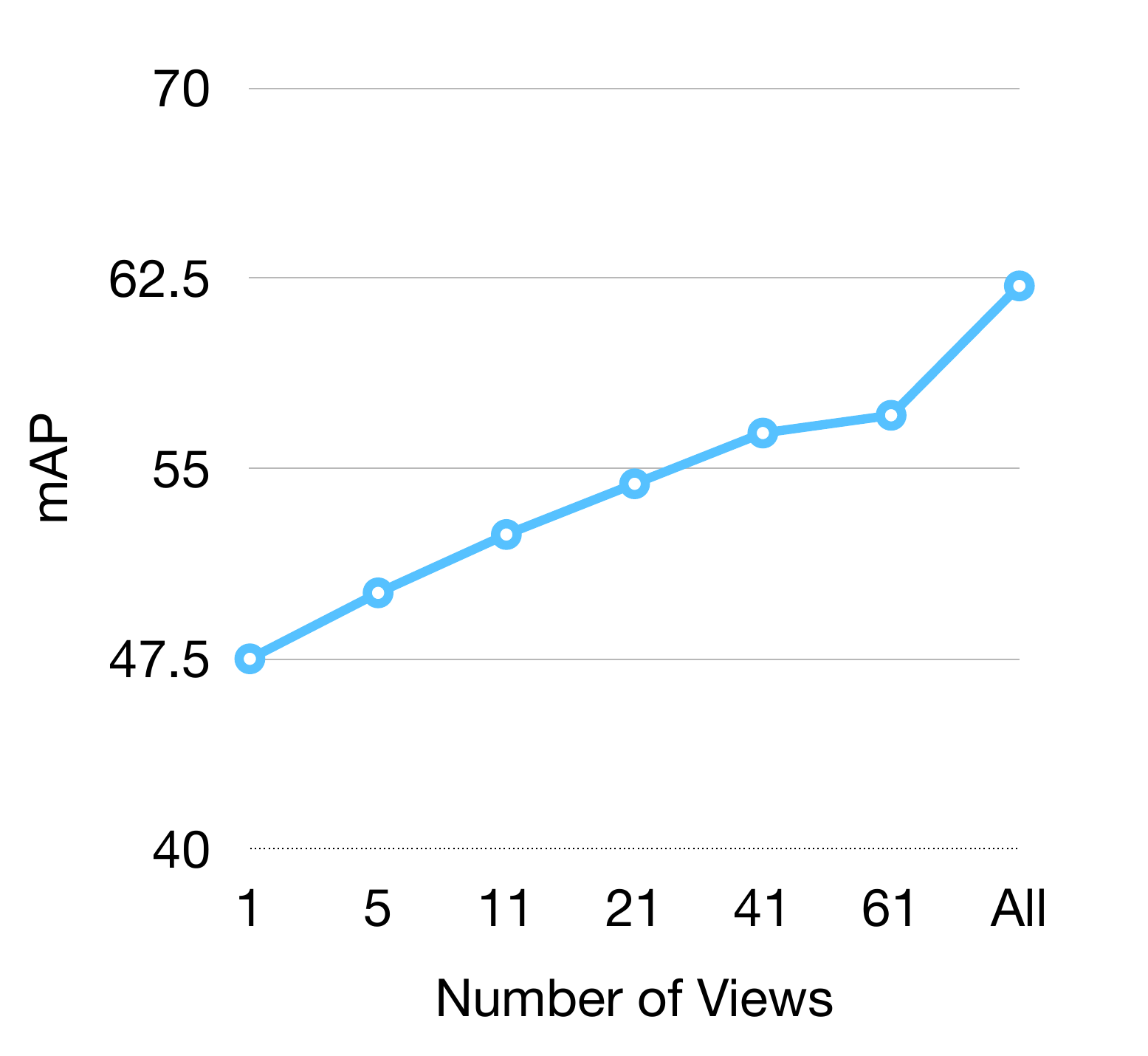}
    \caption{ \textbf{2D mAP Performance Variation with increasing number of context views used}}
     \label{fig:num_views}
\end{figure}

\begin{figure*}[!ht]
     \centering
     \includegraphics[width=\textwidth]{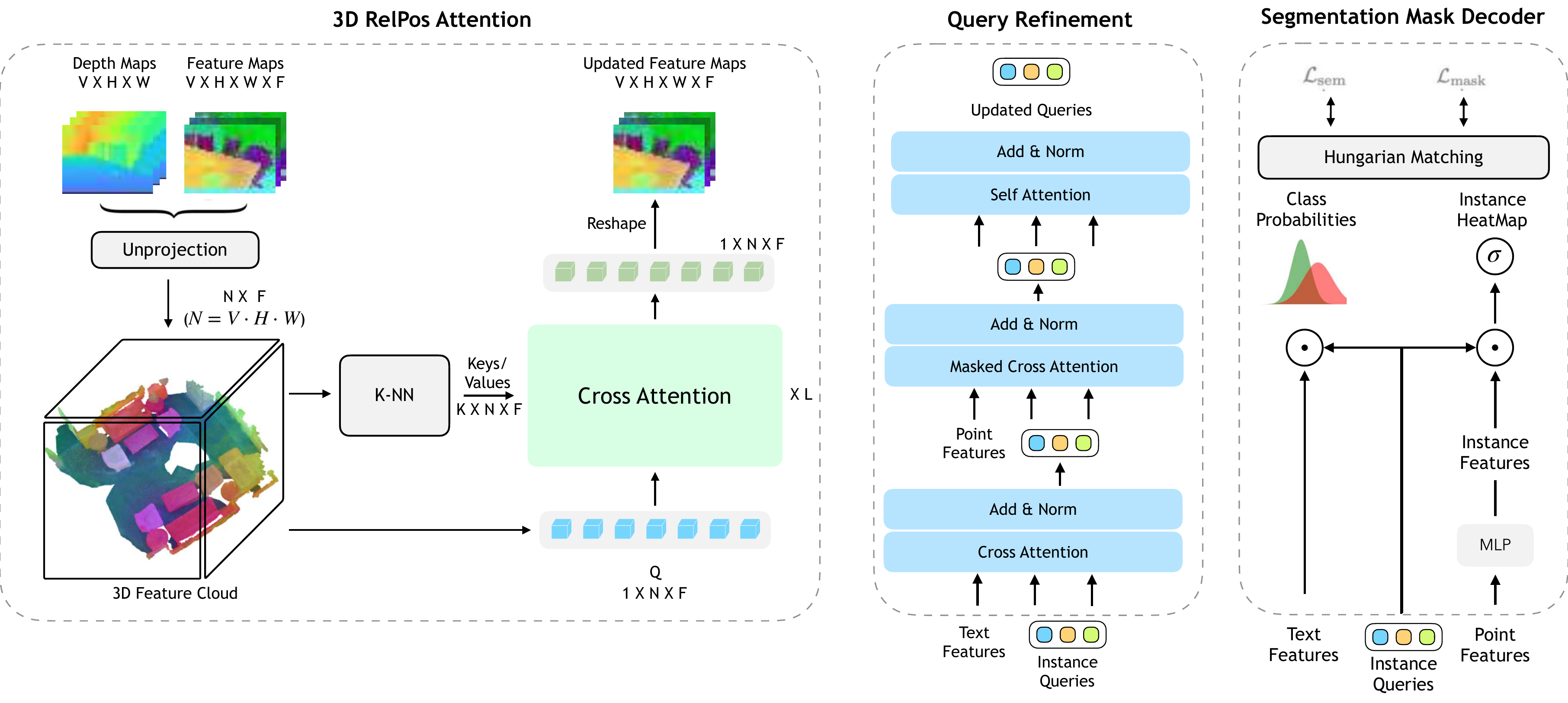}
    \caption{ \textbf{Detailed \model{} Architecture Components}: On the \textbf{Left} is the 3D RelPos Attention module which takes as input the depth, camera parameters and feature maps from all views, lifts the features to 3D to get 3D tokens. Each 3D token serves as a query.  The K-Nearest Neighbors of each 3D token become the corresponding keys and values. The 3D tokens attend to their neighbours for L layers and update themselves. Finally, the 3D tokens are mapped back to the 2D feature map by simply reshaping the 3D feature cloud to 2D multi-view feature maps. On the \textbf{Middle} is the query refinement block where queries first attend to the text tokens, then to the visual tokens and finally undergo self-attention. The text features are optional and are only used in the open-vocabulary decoder setup. On the \textbf{Right} is the segmentation mask decoder head where the queries simply perform a dot-product with visual tokens to decode the segmentation heatmap, which can be thresholded to obtain the segmentation mask. In the Open-Vocabulary decoding setup, the queries also perform a dot-product with text tokens to decode a distribution over individual words. In a closed vocabulary decoding setup, queries simply pass through an MLP to predict a distribution over classes.}
     \label{fig:detailed_arch}
\end{figure*}

\subsection{Variation of Performance with Number of Views}
\label{subsec:num_views}
We examine the influence of the number of views on segmentation performance using the AI2THOR dataset, specifically focusing on the 2D mAP performance metric. The evaluation is conducted by varying the number of \textit{context} images surrounding a given \textit{query} RGB image. Starting from a single-view without any context (N=0), we increment N to 5, 10, 20, 40, 60, and finally consider all images in the scene as context. \model{} takes these $N + 1$ RGB-D images as input, predicts per-pixel instance segmentation for each image, and assesses the 2D mAP performance on the \textit{query} image. The results, depicted in \cref{fig:num_views}, show a continuous increase in 2D mAP with the growing number of views. This observation underscores the advantage of utilizing multiview RGB-D images over single-view RGB images whenever feasible.

\subsection{Inference Time}
We assess the inference time of Mask3D and \model{} by averaging the forward pass time of each model across the entire validation set, utilizing a 40 GB VRAM A100. When fed the mesh-sampled point cloud directly, Mask3D achieves an inference time of 228ms. When provided with the sensor point cloud as input, the inference time increases to 864 ms. Mask3D with sensor point cloud is slower than with mesh point cloud because at the same voxel size (0.02m), more voxels are occupied in sensor point cloud (~110k on avg.) compared to mesh point clouds (~64k on avg.) as mesh-cleaning sometimes discards large portion of the scene. The transfer of features from the sensor point cloud to the mesh point cloud adds an extra 7 ms. \model{}-SwinB, which operates over the sensor point cloud, has an inference time of 960ms.

\section{Additional Implementation Details}
\label{subsec:impl_detail}

The detailed components of our architecture and their descriptions are presented in \cref{fig:detailed_arch}.  

More implementation details are presented below:

\noindent \textbf{Augmentations: } For RGB image augmentation, we implement the Large Scale Jittering Augmentation method from Mask2Former \cite{m2f}, resizing images to a scale between $0.1$ and $2.0$. We adjust intrinsics accordingly post-augmentation and apply color jittering to RGB images. Training involves a consecutive set of $N$ images, typically set to 25. With a 50\% probability, we randomly sample $k$ images from the range $[1, N]$ instead of using all $N$ images. Additionally, instead of consistently sampling $N$ consecutive images, we randomly skip $k$ images in between, where $k$ ranges from 1 to 4.

For 3D augmentations, we adopt the Mask3D \cite{mask3d} approach, applying random 3D rotation, scaling, and jitter noise to the unprojected XYZs. Elastic distortion and random flipping augmentations from Mask3D are omitted due to a slight drop in performance observed in our initial experiments.

\noindent  \textbf{Image Resolutions}
We use a resolution of $256 \times 256$ for ScanNet, $512 \times 512$ for ScanNet200, and AI2THOR. In our AI2THOR experiments, we discovered that employing higher image resolutions enhances the detection of smaller objects, with no noticeable impact on the detection of larger ScanNet-like objects. This observation was confirmed in ScanNet, where we experimented with $512 \times 512$ image resolutions and did not observe any discernible benefit.

\noindent \textbf{Interpolation}
Throughout our model, interpolations are employed in various instances, such as when upsampling the feature map from $1/8$th resolution to $1/4$th. In cases involving depth, we unproject feature maps to 3D and perform trilinear interpolation, as opposed to directly applying bilinear interpolation on the 2D feature maps. For upsampling/downsampling the depth maps, we use the nearest interpolation. Trilinear interpolation proves crucial for obtaining accurate feature maps, particularly at 2D object boundaries like table and floor edges. This is because nearest depth interpolation may capture depth from either the table or the floor. Utilizing trilinear upsampling of feature maps ensures that if the upsampled depth is derived from the floor, it interpolates features from floor points rather than table points.


\noindent  \textbf{Use of Segments: }
Some datasets, such as ScanNet and ScanNet200, provide supervoxelization of the point cloud, commonly referred to as \textit{segments}. Rather than directly segmenting all input points, many 3D methods predict outputs over these segments. Specifically, Mask3D \cite{mask3d} featurizes the input points and then conducts mean pooling over the features of points belonging to a segment, resulting in one feature per segment. Following prior work, we also leverage segments in a similar manner. We observe that utilizing segments is crucial for achieving good mAP performance, while it has no discernible impact on mAP25 performance. We suspect that this phenomenon may arise from the annotation process of these datasets. Humans were tasked with labelling segments rather than individual points, ensuring that all points within a segment share the same label. Utilizing segments with our models guarantees that the entire segment is labelled with the same class. It's worth noting that in AI2THOR, our method and the baselines do not utilize these segments, as they are not available.


\noindent  \textbf{Post-hoc output transfer vs feature transfer: } \model{} takes the sensor point cloud as input and generates segmentation output on the benchmark-provided point cloud. In this process, we featurize the sensor point cloud and transfer these features from the sensor point cloud to the benchmark-provided point cloud. Subsequently, we predict segmentation outputs on this benchmark-provided feature cloud and supervise the model with the labels provided in the dataset. An alternative approach involves segmenting and supervising the sensor RGB-D point cloud and later transferring the segmentation output to the benchmark point cloud for evaluation. We experimented with both strategies and found them to yield similar results. However, as many datasets provide segmentation outputs only on the point cloud, transferring labels to RGB-D images for the latter strategy requires careful consideration. This is due to the sparser nature of the provided point cloud compared to the RGB-D sensor point cloud, and factors such as depth noise and misalignments can contribute to low-quality label transfer. Consequently, we opt for the former strategy in all our experiments.

\noindent  \textbf{Depth Hole-Infilling: } The sensor-collected depth maps usually have holes around object boundaries and shiny/transparent surfaces. We perform simple OpenCV depth inpainting to fill these holes. We tried using neural-based depth completion methods and NERF depth-inpainting but did not observe significant benefits. 

\noindent  \textbf{AI2THOR Data Collection: } AI2THOR \cite{ai2thor} is an embodied simulator where an agent can navigate within a house, execute actions, and capture RGB-D images of the scene. We load the structurally generated houses from ProcTHOR \cite{procthor} into the AI2THOR simulator, and place an agent randomly at a navigable point provided by the simulator. The agent performs a single random rotation around its initial location and captures an RGB-D frame. This process is repeated, with the agent spawning at another random location, until either all navigable points are exhausted or a maximum of $N = 120$ frames is collected. While ProcTHOR offers 10,000 scenes, we randomly select only 1,500 scenes to match the size of ScanNet. Additionally, we retain scenes with fewer than $100$ objects, as our model utilizes a maximum of $100$ object queries.

\begin{figure*}[!t]
\centering
    \includegraphics[width=0.95\textwidth]{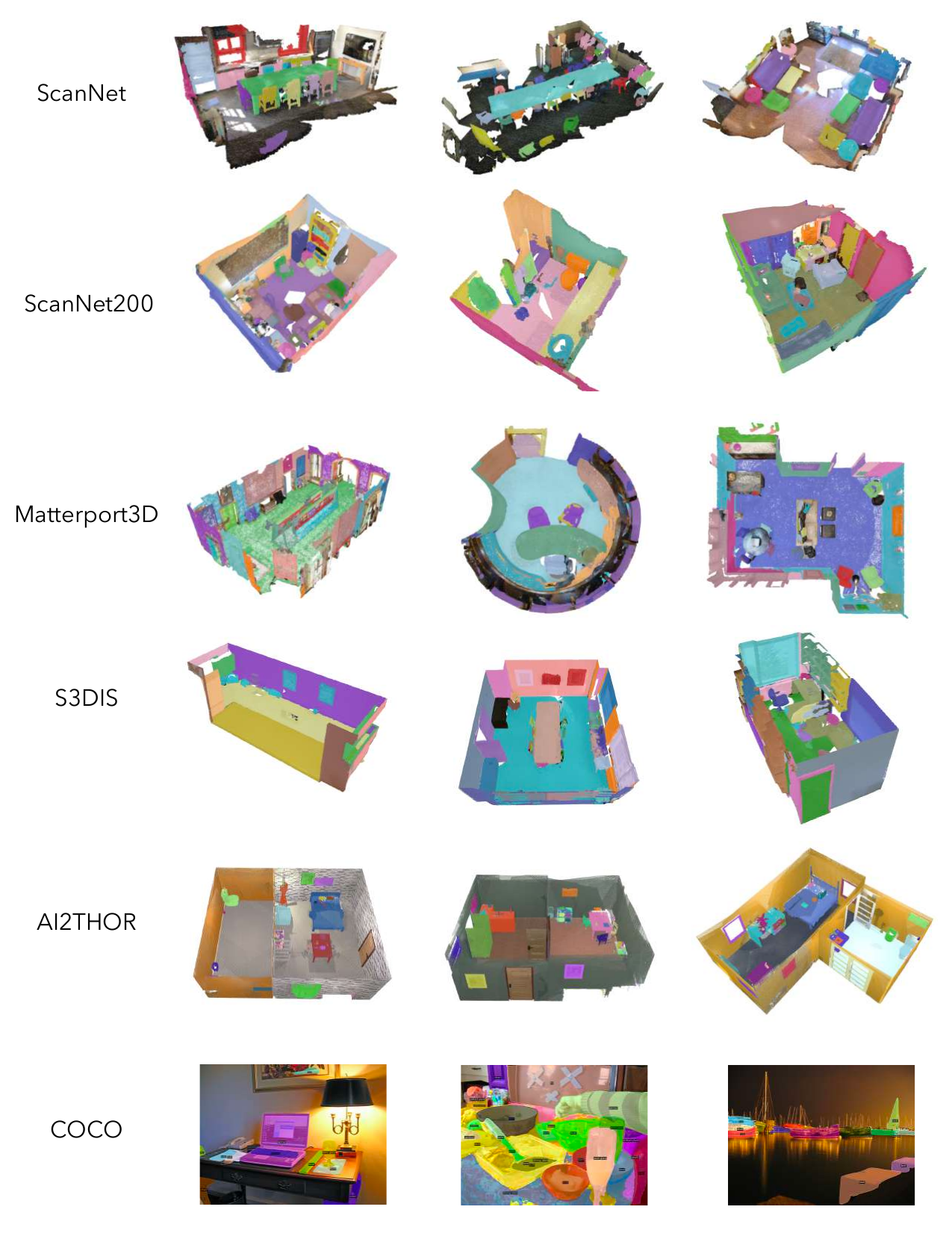}
    \caption{Qualitative Results on various 3D and 2D datasets}
    \label{fig:supp_qual}
\end{figure*}


\section{Qualitative Results}
\cref{fig:supp_qual} shows qualitative visualizations of \model{} for various 3D and 2D datasets.

%% file: tables/test_real_in_domain_3d.tex
\begin{table*}
\caption{\textbf{Evaluation on Test Set of Established 3D Benchmarks.}}
\label{tab:test_established_3d}
\begin{subtable}[t]{0.51\textwidth}
\small
\centering
\caption{\textbf{Comparison on ScanNet for Instance Segmentation Task.}}
\label{tab:test_scannet_ins}
\resizebox{\textwidth}{!}{%
\begin{tabular}{llccccc}
\midrule
Input & Model & mAP &  mAP50 & mAP25 \\ \midrule
\multirow{2}{1.9cm}{Sensor RGBD Point Cloud} 
 & \model{}-Swin-B (Ours) & \textbf{47.7} & \textbf{71.2} & \textbf{86.2}  \\ \\ \midrule
\multirow{5}{1.9cm}{Mesh Sampled Point Cloud} & SoftGroup \cite{softgroup} & 50.4 & 76.1 & 86.5 \\
& PBNet \cite{pbnet} & 57.3 & 74.7  &  82.5 \\
& Mask3D \cite{mask3d} & 56.6 & 78.0 & 87.0 \\
& QueryFormer \cite{queryformer} & 58.3 & \textbf{78.7} & \textbf{87.4} \\
& MAFT \cite{maft} & \textbf{59.6} & 78.6 & 86.0\\ 
 \midrule
\end{tabular}
}
\end{subtable}
\hspace{\fill}
\begin{subtable}[t]{0.48\textwidth}
\small
\centering
\caption{\textbf{Comparison on ScanNet for Semantic Segmentation Task.}}
\label{tab:test_scannet_sem}
\begin{tabular}{llc} 
\midrule
Input & Model & mIoU \\ \midrule
\multirow{5}{2.2cm}{Sensor RGBD Point Cloud} 
& MVPNet \cite{mvpnet} & 64.1 \\
& BPNet \cite{bpnet} & \textbf{74.9} \\
& DeepViewAgg \cite{deepview} & - \\
 & \model{}-Swin-B (Ours) & 74.4 \\ \midrule
\multirow{1}{2.2cm}{Rendered RGBD Point Cloud} & VMVF \cite{virtualview} & \textbf{74.6} \\ \\\midrule
\multirow{4}{2.2cm}{Mesh Sampled Point Cloud} & Point Transformer v2 \cite{ptv2} & 75.2 \\ 
& Stratified Transformer \cite{stratified} & 74.7 \\
& OctFormer \cite{octformer} & 76.6 \\
& Swin3D-L \cite{swin3d} & \textbf{77.9}  \\
 \midrule
Zero-Shot & OpenScene \cite{openscene} & - \\
\midrule
\end{tabular} 
\end{subtable}
\bigskip \\
\begin{subtable}[t]{0.50\textwidth}
\vspace*{-0.3 cm}
\small
\centering
\caption{\textbf{Comparison on ScanNet200 for Instance Segmentation Task.}}
\label{tab:test_scannet200_ins}
\resizebox{\textwidth}{!}{
\begin{tabular}{llccc}
\midrule
 & Model  & mAP &  mAP50 & mAP25 \\ \midrule
 \multirow{2}{1.9cm}{Sensor RGBD Point Cloud} 
 & \model{}-Swin-B (Ours) &  \textbf{27.2}        &  \textbf{39.4}         &  \textbf{47.5}   \\ \\ \midrule
\multirow{3}{1.9cm}{Mesh Sampled Point Cloud} & Mask3D \cite{mask3d} & \textbf{27.8} & \textbf{38.8}  & \textbf{44.5} \\
& QueryFormer \cite{queryformer} & - & - & - \\
& MAFT \cite{maft} & - & - & - \\ 
 \midrule
 Zero-Shot & OpenMask3D \cite{openmask3d} & - & - & - \\
 \midrule
\end{tabular}
}
\end{subtable}
\hspace{\fill}
\begin{subtable}[t]{0.48\textwidth}
\small
\centering
\caption{\textbf{Comparison on ScanNet200 for Semantic Segmentation Task.}}
\label{tab:test_scannet200_sem}

\begin{tabular}{llc}
\midrule
Input & Model  & mIoU \\ \midrule
\multirow{2}{2cm}{Sensor RGBD Point Cloud} 
 & \model{}-Swin-B (Ours) &  \textbf{36.8}  \\ \\ \midrule
\multirow{3}{2cm}{Mesh Sampled Point Cloud} & LGround \cite{scannet200} & 27.2 \\
& CeCo \cite{ceco} & \textbf{34.0} \\
& Octformer \cite{octformer} & 32.6 \\
 \midrule
\end{tabular}
\end{subtable}

\end{table*}

%% file: 7_s3dis_m3d.tex
\subsection{Evaluation on S3DIS and Matterport3D}
\label{subsec:m3d_s3dis}
We also benchmark \model{} on Matterport3D \cite{matterport} and S3DIS \cite{s3dis} datasets.

\textbf{Matterport:} Matterport3D comprises 90 building-scale scenes, further divided into individual rooms, with 1554 training rooms and 234 validation rooms. The dataset provides a mapping from each room to the camera IDs that captured images for that room. After discarding 158 training rooms and 18 validation rooms without a valid camera mapping, we are left with 1396 training rooms and 158 validation rooms. For instance segmentation results, we train the state-of-the-art Mask3D \cite{mask3d} model on the same data (reduced set after discarding invalid rooms). For semantic segmentation, we conduct training and testing on the reduced set, while baseline numbers are taken from the OpenScene \cite{openscene} paper, trained and tested on the original data. Given the small size of the discarded data, we do not anticipate significant performance differences. The official benchmark of Matterport3D tests on 21 classes; however, OpenScene also evaluates on 160 classes to compare with state-of-the-art models on long-tail distributions. We follow them and report results in both settings.


\textbf{S3DIS: } S3DIS comprises 6 building-scale scenes, typically divided into 5 for training and 1 for testing. The dataset provides raw RGB-D images, captured panorama images, and images rendered from the mesh obtained after reconstructing the original sensor data. Unlike Matterport3D, S3DIS do not provide undistorted raw images; thus, we use the provided rendered RGB-D images. Some rooms in S3DIS have major misalignments between RGB-D images and point clouds, which we partially address by incorporating fixes from DeepViewAgg \cite{deepview} and introducing our own adjustments. Despite these fixes, certain scenes still exhibit significantly low overlap between RGB-D images and the provided mesh-sampled point cloud. To mitigate this, we query images from other rooms and verify their overlap with the provided point cloud for a room. This partially helps in addressing the low overlap issue.

The official S3DIS benchmark evaluates 13 classes. Due to the dataset's small size, some models pre-train on additional datasets like ScanNet, as seen in SoftGroup \cite{softgroup}, and on Structured3D datasets \cite{structured3d}, consisting of 21,835 rooms, as done by Swin3D-L \cite{swin3d}. Similar to Mask3D \cite{mask3d}, we report results in both settings of training from scratch and starting from weights trained on ScanNet. 

Like ScanNet and ScanNet200, both S3DIS and Matterport3D undergo post-processing of collected RGB-D data to construct a mesh, from which a point cloud is sampled and labeled. Hence, we train both Mask3D \cite{mask3d} and our model using RGB-D sensor point cloud data and evaluate on the benchmark-provided point cloud. Additionally, we explore model variants by training and testing them on the mesh-sampled point cloud for comparative analysis.

The results are shown in \cref{tab:matterport_s3dis}. We draw the following conclusions:

\input{tables/matterport_s3dis}
\noindent \model{} outperforms SOTA 3D models on Matterport3D Instance Segmentation Benchmark across all settings (\cref{tab:matterport_ins}) 


\noindent \model{} sets a new state-of-the-art on Matterport3D Semantic Segmentation Benchmark (\cref{tab:matterport_sem}): Our model achieves superior performance in both the 21 and 160 class settings. It also largely outperforms OpenScene \cite{openscene} on both settings. OpenScene is a zero-shot method while \model{} is supervised in-domain, making this comparison unfair. However, OpenScene notes that their zero-shot model outperforms fully-supervised models in 160 class setup as their model is robust to rare classes while the supervised models can severely suffer in segmenting long-tail. ConceptFusion \cite{conceptfusion}, another open-vocabulary 3D segmentation model, also draws a similar conclusion. With this result, we point to a possibility of supervising in 3D while also being robust to long-tail by simply utilizing the strong 2D pre-trained weight initialization. 


On S3DIS Instance Segmentation Benchmark (\cref{tab:s3dis_ins}), in the setup where baseline Mask3D start from ScanNet pre-trained checkpoint, our model outperforms them in the RGBD point cloud setup but obtains lower performance compared to mesh sampled point cloud methods and when compared on the setup where all models train from scratch.

On S3DIS Semantic Segmentation Benchmark (\cref{tab:s3dis_sem}, \model{} trained with ScanNet weight initialization outperforms all RGBD point cloud based methods, while achieving competitive performance on mesh sampled point cloud. When trained from scratch, it is much worse than other baselines. Given the limited dataset size of S3DIS with only 200 training scenes, we observe severe overfitting. 

%% file: tables/matterport_s3dis.tex
\begin{table*}

\caption{\textbf{Evaluation on Matterport3D \cite{matterport} and S3DIS \cite{s3dis} datasets.}}
\label{tab:matterport_s3dis}
\begin{subtable}[t]{0.50\textwidth}
\small
\centering
\caption{\textbf{Comparison on Matterport3D for Instance Segmentation Task.}}
\label{tab:matterport_ins}
\resizebox{\textwidth}{!}{%
\begin{tabular}{llcccc}
\midrule
 & & \multicolumn{2}{c}{\textbf{21}} & \multicolumn{2}{c}{\textbf{160}} \\
  \cmidrule(lr){3-4} \cmidrule(lr){5-6}
Input & Model & mAP & mAP25 & mAP & mAP25 \\ \midrule
\multirow{3}{1.9cm}{Sensor RGBD Point Cloud} & Mask3D \cite{mask3d} & 7.2 & 16.8 & 2.5 & 10.9  \\
 & \model{}-ResNet50 (Ours) & 22.5 & 56.4 & 11.5  & 27.6  \\
 & \model{}-Swin-B (Ours) & \textbf{24.7} & \textbf{63.8} & \textbf{14.5} &  \textbf{36.8} \\ \midrule
\multirow{2}{1.9cm}{Mesh Sampled Point Cloud} & 
 Mask3D \cite{mask3d} & \textbf{22.9} & \textbf{55.9} & \textbf{11.3} & \textbf{23.9} \\ \\
 \midrule
\end{tabular}
}
\end{subtable}
\hspace{\fill}
\begin{subtable}[t]{0.50\textwidth}
\small
\centering
\caption{\textbf{Comparison on Matterport3D for Semantic Segmentation Task.}}
\label{tab:matterport_sem}
\resizebox{\textwidth}{!}{%
\begin{tabular}{llcccc} 
\midrule
 & & \multicolumn{2}{c}{\textbf{21}} & \multicolumn{2}{c}{\textbf{160}} \\
  \cmidrule(lr){3-4} \cmidrule(lr){5-6}
Input & Model & mIoU & mAcc & mIoU & mAcc \\ \midrule
\multirow{2}{2.1cm}{Sensor RGBD Point Cloud} 
 & \model{}-ResNet50 (Ours) & 54.5 & 65.8 & 22.4 & 28.5   \\
 & \model{}-Swin-B (Ours) & \textbf{57.3} & \textbf{69.4} & \textbf{28.6} & \textbf{38.2} \\ \midrule
\multirow{3}{2.2cm}{Mesh Sampled Point Cloud} &  TextureNet \cite{texturenet}  & - & 63.0 & - & - \\
& DCM-Net \cite{dcmnet} & - & \textbf{67.2} & - & - \\
& MinkowskiNet \cite{minkowski} & \textbf{54.2} & 64.6 & - & \textbf{18.4} \\
\midrule
Zero-Shot & OpenScene \cite{openscene} & \textbf{42.6} & \textbf{59.2} & - & \textbf{23.1} \\
\midrule
\end{tabular} 
}
\end{subtable}
\bigskip \\
\begin{subtable}[t]{0.50\textwidth}
\vspace*{-0.5 cm}
\small
\centering
\caption{\textbf{Comparison on S3DIS Area5 for Instance Segmentation Task.} ($^\dagger$ = uses additional data)}
\label{tab:s3dis_ins}
\resizebox{\textwidth}{!}{
\begin{tabular}{llccc}
\midrule
 & Model  & mAP &  mAP50 & mAP25 \\ \midrule
\multirow{5}{1.9cm}{RGBD Point Cloud} & Mask3D \cite{mask3d} & 40.7 & 54.6 & 64.2  \\
& Mask3D \cite{mask3d} $^{\dagger}$ & 41.3 & 55.9 & 66.1  \\
 & \model{}-ResNet50 (Ours) & 36.3 & 48.0 & 61.2  \\
 & \model{}-ResNet50 $^{\dagger}$ (Ours) & \textbf{44.7} & \textbf{57.7} & 67.5  \\
 & \model{}-Swin-B $^{\dagger}$ (Ours) &  43.0       &  56.4         &  \textbf{70.0}   \\ \midrule
\multirow{5}{1.9cm}{Mesh Sampled Point Cloud} 
& SoftGroup \cite{softgroup} $^{\dagger}$ & 51.6 & 66.1 & - \\
& Mask3D \cite{mask3d} & 56.6 & 68.4  & 75.2 \\
& Mask3D \cite{mask3d} $^{\dagger}$ & \textbf{57.8} & \textbf{71.9}  & \textbf{77.2} \\
& QueryFormer \cite{queryformer} & 57.7 & 69.9 & - \\
& MAFT \cite{maft} & - & 69.1 & 75.7 \\ 
 \midrule
\end{tabular}
}
\end{subtable}
\hspace{\fill}
\begin{subtable}[t]{0.42\textwidth}
\vspace*{-0.5 cm}
\small
\centering
\caption{\textbf{Comparison on S3DIS for Semantic Segmentation Task.} ($^\dagger$ = uses additional data)}
\label{tab:s3dis_sem}
\resizebox{\textwidth}{!}{%

\begin{tabular}{llc} 
\midrule
Input & Model & mIoU \\ \midrule
\multirow{6}{2.2cm}{RGBD Point Cloud} 
& MVPNet \cite{mvpnet} & 62.4 \\
  & VMVF \cite{virtualview} & 65.4 \\ 
& DeepViewAgg \cite{deepview} & 67.2 \\
 & \model{}-ResNet50 (Ours) & 59.7  \\
  & \model{}-ResNet50 $^{\dagger}$ (Ours) & 66.8 \\
 & \model{}-Swin-B $^{\dagger}$ (Ours) & \textbf{68.6} \\ \midrule

\multirow{3}{2.2cm}{Mesh Sampled Point Cloud} & Point Transformer v2 \cite{ptv2} & 71.6 \\ 
& Stratified Transformer \cite{stratified} & 72.0 \\
& Swin3D-L \cite{swin3d} $^{\dagger}$ & \textbf{74.5}  \\
 \midrule
\end{tabular} 
}
\end{subtable}

\end{table*}

%% file: tables/scannet200_detailed.tex
\begin{table*}
\caption{\textbf{Detailed ScanNet200 results for Instance Segmentation} ($^\mathsection$ = trained by us using official codebase)}
\label{tab:scannet200_detailed}
\resizebox{\textwidth}{!}{%

\begin{tabular}{llcccccccccccc}
\toprule
Input & Model & \multicolumn{3}{c}{All} &  \multicolumn{3}{c}{Head} & \multicolumn{3}{c}{Common} & \multicolumn{3}{c}{Tail} \\
 \cmidrule(lr){3-5} \cmidrule(lr){6-8} \cmidrule(lr){9-11} \cmidrule(lr){12-14}  
& & mAP &  mAP50 & mAP25 & mAP &  mAP50 & mAP25 & mAP &  mAP50 & mAP25 & mAP &  mAP50 & mAP25\\ \midrule
\multirow{3}{*}{Sensor RGBD point cloud} & Mask3D $^{\mathsection}$ \cite{mask3d} & 15.5 & 21.4 & 24.3 & 21.9 & 31.4 &  37.1 & 13.0 & 17.2 & 18.9 & 7.9 & 10.3 & 11.5 \\
 & \model{}-ResNet50 (Ours) & 25.6 & 36.9 & 43.8 & 34.8 & 51.1 & 63.9 & 23.4 & 33.4  & 37.9 & 17.8 & 24.9 & 28.1 \\
 & \model{}-Swin-B (Ours) & \textbf{31.5}    &   \textbf{45.3}      & \textbf{53.1}  & 37.5 &         54.2    &      \textbf{66.1}  &  \textbf{31.6}     &     \textbf{43.9} &         \textbf{50.2} & \textbf{24.1}       &   \textbf{36.6}      &    \textbf{41.2}  \\ \midrule
\multirow{1}{*}{Mesh Sampled point cloud} & Mask3D \cite{mask3d} & 27.4 & 37.0  & 42.3 & \textbf{40.3} & \textbf{55.0} & 62.2 & 22.4 & 30.6 & 35.4 & 18.2 & 23.2 & 27.0\\
 \bottomrule
\end{tabular}
}
\end{table*}